\documentclass{article}
\usepackage{iclr,times}
\usepackage{caption}
\usepackage[hidelinks]{hyperref}
\usepackage{url}
\usepackage{subcaption}
\usepackage{graphicx}
\usepackage{enumitem}
\usepackage{booktabs}
\usepackage{amsmath}

\title{Twitch plays pokemon, machine learns twitch: Unsupervised Context-Aware Anomaly Detection for Identifying Trolls in Streaming Data }

\author{Albert Haque\\
Department of Computer Science\thanks{Completed as a class project for CS 363D: Data Mining at the University of Texas at Austin in May 2014. Dataset and code is available at: \url{https://github.com/ahaque/twitch-troll-detection}}\\
University of Texas at Austin\\
Austin, TX 78712, USA
}

\iclrfinalcopy
\begin{document}

\maketitle

\begin{abstract}
With the increasing importance of online communities, discussion forums, and customer reviews, Internet ``trolls" have proliferated thereby making it difficult for information seekers to find relevant and correct information. In this paper, we consider the problem of detecting and identifying Internet trolls, almost all of which are human agents. Identifying a human agent among a human population presents significant challenges compared to detecting automated spam or computerized robots.  To learn a troll's behavior, we use contextual anomaly detection to profile each chat user. Using clustering and distance-based methods, we use contextual data such as the group's current goal, the current time, and the username to classify each point as an anomaly. A user whose features significantly differ from the norm will be classified as a troll. We collected 38 million data points from the viral Internet fad, Twitch Plays Pokemon. Using clustering and distance-based methods, we develop heuristics for identifying trolls. Using MapReduce techniques for preprocessing and user profiling, we are able to classify trolls based on 10 features extracted from a user's lifetime history.
\end{abstract}

\section{Introduction}

In Internet slang, a ``troll" is a person who sows discord on the Internet by starting arguments or upsetting people, by posting inflammatory, extraneous, or off-topic messages in an online community (such as a forum, chat room, or blog), either accidentally or with the deliberate intent of provoking readers into an emotional response or of otherwise disrupting normal on-topic discussion \citep{wikipedia}.

Application of the term troll is subjective. Some readers may characterize a post as trolling, while others may regard the same post as a legitimate contribution to the discussion, even if controversial \citep{wikipedia}. Regardless of the circumstances, controversial posts may attract a particularly strong response from those unfamiliar with the dialogue found in some online, rather than physical, communities \citep{wikipedia}. Experienced participants in online forums know that the most effective way to discourage a troll is usually to ignore it, because responding tends to encourage trolls to continue disruptive posts – hence the often-seen warning: ``Please don't feed the trolls" \citep{heilmann2012web}.

Given the novelty of the Twitch Plays Pokemon event, we wanted to leverage known clustering heuristics to identify trolls in real-time. This requires the application of several preprocessing steps which we performed in MapReduce \citep{dean2008mapreduce}, a distributed processing system.

The contributions of this project are as follows. First, we propose a set of features used for identifying trolls in the viral, croudsourced Internet game, Twitch Plays Pokemon \citep{tpp}. We use context-based techniques to understand the scenario a human is faced when entering input into the chatroom. We then compare the effects of different distance measures to understand their strengths and weaknesses.

The second major contribution of this paper is an online classification algorithm. It is initially trained using unsupervised methods on offline data. When switched to online mode, this algorithm updates as new data points are received from a live stream. The primary purpose of this project, through these two contributions, is to in realtime, distinguish between trolls and humans on the Internet. Future work can be extended to email classifiers, comment moderation, and anomaly detection for online forums, reviews, and other communities. The full dataset and code are made available online.

The remainder of this paper is organized as follows: In Section \ref{sec:background} we give an overview of Twitch Plays Pokemon and discuss the dataset we collected. Section \ref{sec:features} outlines how we extract features from our dataset. These features are then used by algorithms outlined in Section \ref{sec:algorithms}. Results are shown in Section \ref{sec:algorithms} followed by a discussion in Section \ref{sec:discussion}. We then conclude in Section \ref{sec:conclusion} by discussing related and future work.

\section{Background}\label{sec:background}
\subsection{Twitch Plays Pokemon}
\begin{figure}[t]
\centering
\includegraphics[width=0.8\textwidth]{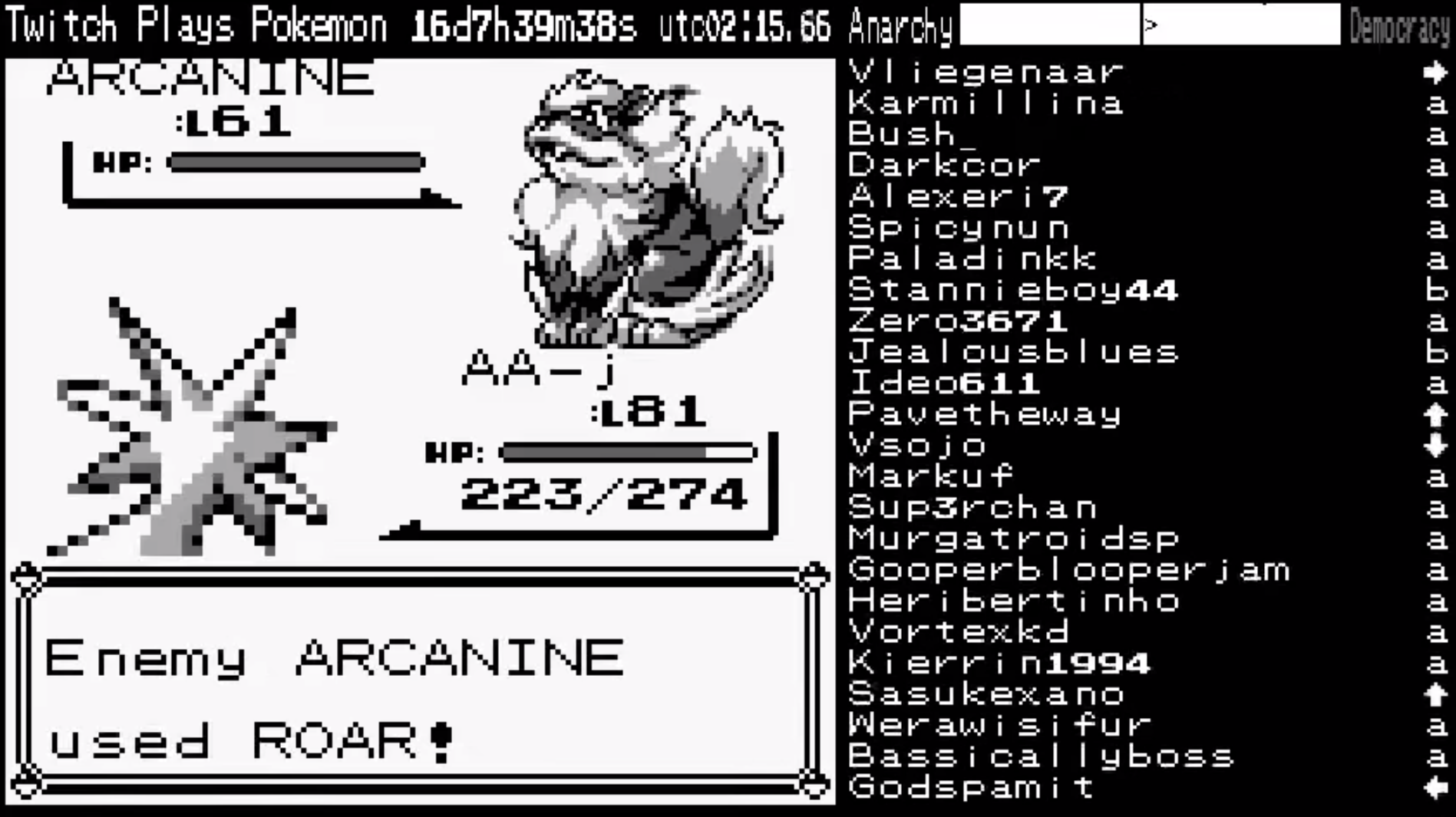}
\caption{Screenshot of Twitch Plays Pokemon. User input messages are shown on the right side. The top of the list is the front of the queue.}
\vspace{-2 mm}
\label{fig:tpp_screenshot}
\end{figure}

Twitch Plays Pokemon is a ``social experiment" and channel on the video streaming website Twitch.tv, consisting of a crowd-sourced attempt to play Game Freak and Nintendo's Pokemon video games by parsing commands sent by users through the channel's chat room. Users can input any message into the chat but only the following commands are recognized by the bot (server-side script parsing inputs): up, down, left, right, start, select, a, b, anarchy, democracy. Any other message will have no effect on the game. This is shown in Figure \ref{fig:tpp_screenshot}.

Anarchy and democracy refer to the ``mode" of the game. In anarchy mode, inputs are executed in pseudo-FIFO order (see next paragraph for definition). In democracy mode, the bot collects user input for 20 seconds, after which it executes the most frequently entered command. The mode of the game changes when a certain percentage of users vote for democracy or anarchy. The mode will remain in that state until users vote the other direction.

In anarchy mode, the bot attempts to execute commands sequentially in a pseudo-FIFO order. Many commands are skipped to empty the queue faster and thus an element of randomness is introduced, hence the name pseudo-FIFO. The uncertainty in the queue reduces to selecting an action at random where the frequency of user input serves as the underlying probability distribution.

\subsection{Dataset}

We wrote an Internet Relay Chat robot to collect user messages, commands, time, and user information. The bot collected data for 73 days. It is 3.4 GB in size and contains approximately 38 million data points. A data point is defined as a user's chat message and metadata such as their username and timestamp at millisecond resolution. Figure \ref{fig:raw_data_point} shows a data point in an XML-like format.

\begin{figure}[ht]
\centering
$<$date$>$2014-02-14$<$/date$>$$<$time$>$08:16:23$<$/time$>$$<$user$>$yeniuss$<$/user$>$$<$msg$>$A$<$/msg$>$
\caption{XML-Like Data Point}
\label{fig:raw_data_point}
\end{figure}

\begin{figure}[h]
\centering
\includegraphics[width=0.8\textwidth]{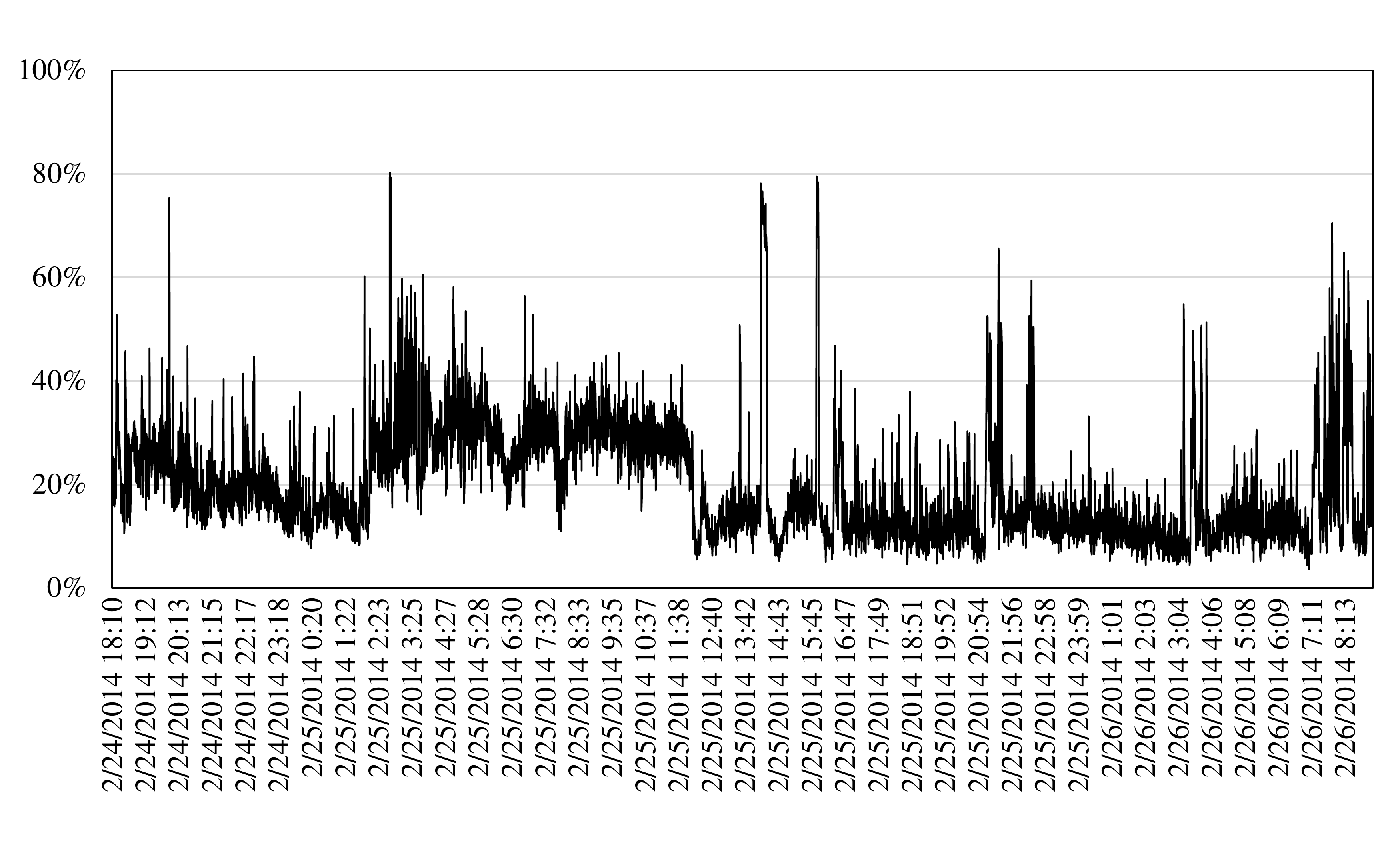}
\vspace{-4 mm}
\caption{Percent of Messages Considered Spam. Each point corresponds to a 20-period moving average with periods of one second each.}
\label{fig:percent_messages_spam}
\end{figure}

Users can enter non-commands into the chat as well. We call these types of messages ``spam." It is possible for a spam message to contribute positively to a conversation. However, these messages have no impact on the game. When a major milestone is accomplished in the game, it is common to see an increase in the frequency of spam messages. Certain spikes can be seen in Figure \ref{fig:percent_messages_spam}. It is possible to gain further insight about each spam message through natural language processing techniques but that is outside the scope of this project.

Since a human logs into the system with the same username, we are able to build profiles of each user and learn from their past messages and commands. We do this by constructing a ``context" every twenty seconds using all messages submitted during the twenty seconds. The notion of contexts is explained in the next section.

\section{Feature Selection}\label{sec:features}
\subsection{Context}
We define a context as a summary of a period of time. It can also be thought of as a sliding window of time with a specific duration. The context stores important information such as the button frequencies, number of messages, and percentage of spam messages.

The context is critical to distinguishing trolls from non-trolls. In Pokemon, there are times when a specific set of button inputs must be entered before the game can proceed. If one button is entered out of sequence, the sequence must restart. For those familiar with the Pokemon game, we are referring to the ``cut" sequence where the player must cut a bush. Sequence 1 (Figure \ref{fig:sequence1}) below lists the required button chain.

\begin{figure}[h]
\centering
DOWN $>$ START $>$ UP $>$ UP $>$ A $>$ UP $>$ A $>$ DOWN $>$ DOWN $>$ A
\caption{Sequence 1. An ordered list of button commands.}
\label{fig:sequence1}
\end{figure}

In normal gameplay, the START button brings up the menu and delays progress of the game. Users entering START are not contributing to the group's goal and are then labeled as trolls. However, if the collective goal is to cut a bush or execute Sequence 1, users inputting START may actually be non-troll users. Therefore we need to distinguish between Sequence 1 and normal gameplay (where START is not needed).

This is done using a context. Every context maintains statistics about all user inputs during a period of time (context duration). If many people are pressing START -- more than usual -- it is possible we are in a Sequence 1 event and we can classify accordingly.

\begin{figure}[h]
\centering
\vspace{-22 mm}
\begin{subfigure}[b]{0.49\textwidth}
\includegraphics[
page=1,
width=\textwidth,
height=\textheight,
keepaspectratio,
trim=55 180 55 20pt,
]{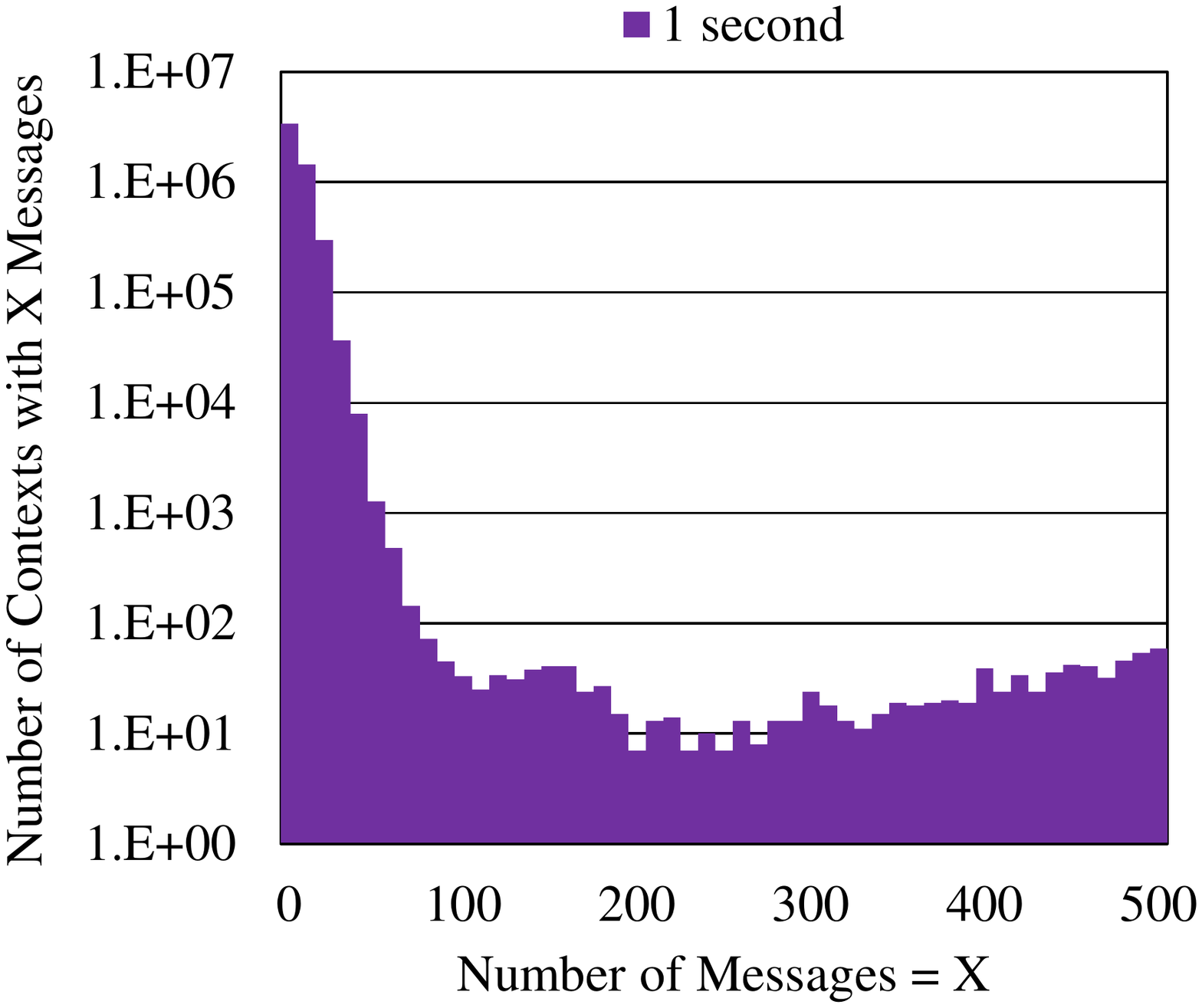}
\end{subfigure}
\begin{subfigure}[b]{0.49\textwidth}
\includegraphics[
page=1,
width=\textwidth,
height=\textheight,
keepaspectratio,
trim=55 180 55 20pt,
]{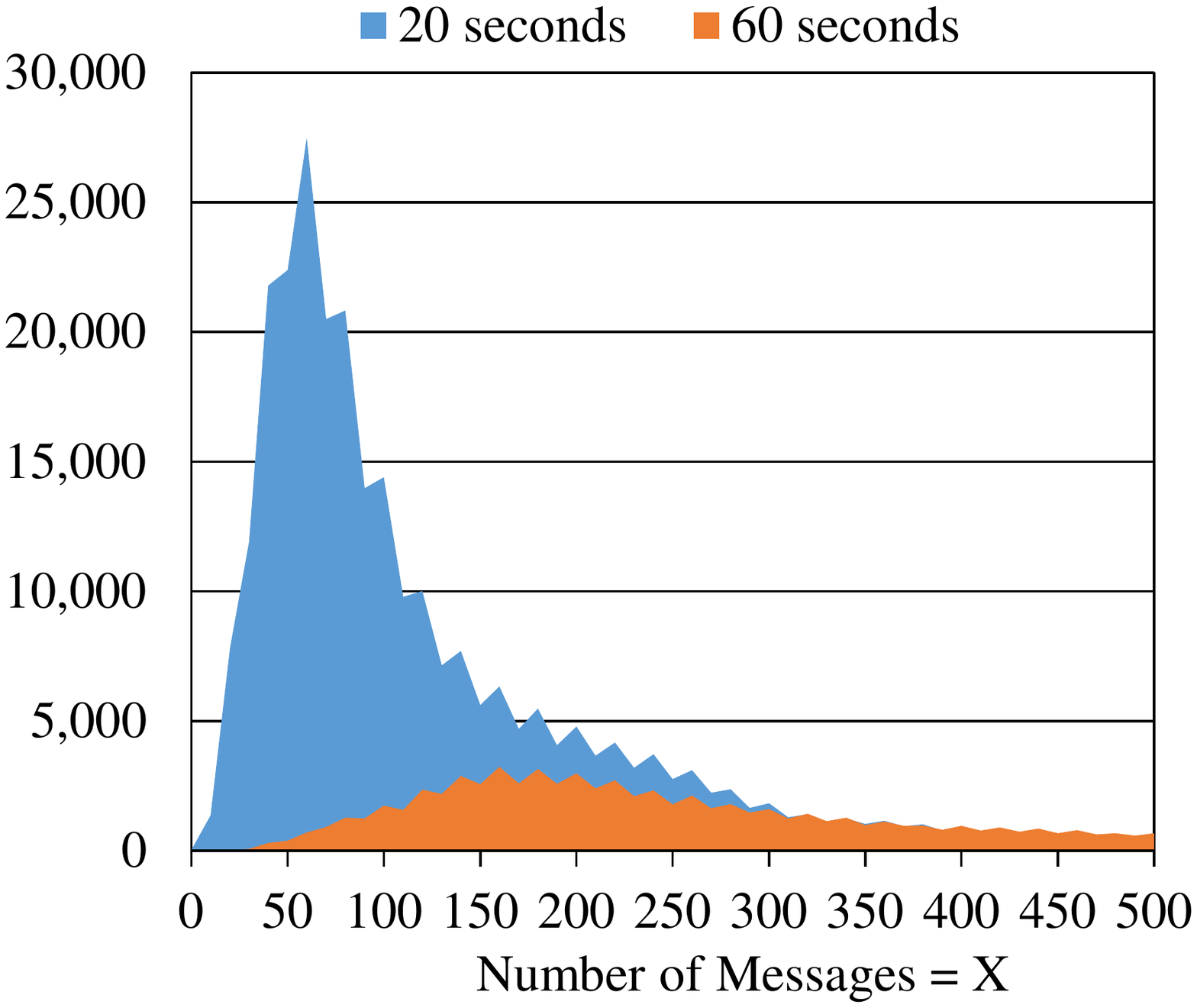}
\end{subfigure}
\caption{Effect of Varying the Context Duration on Messages per Contex}
\label{fig:context_size_historgram}
\end{figure}

We experimented with context durations of 60 seconds, 20 seconds, and 1 second. Every 60, 20, and 1 second(s), we analyze the chat log create a single context. As the context duration increases, the effect of noise is reduced (due to less granular samples as shown in Figure \ref{fig:context_size_historgram}), and the context is smoother.

\begin{figure}[h]
\centering
\includegraphics[width=0.8\textwidth]{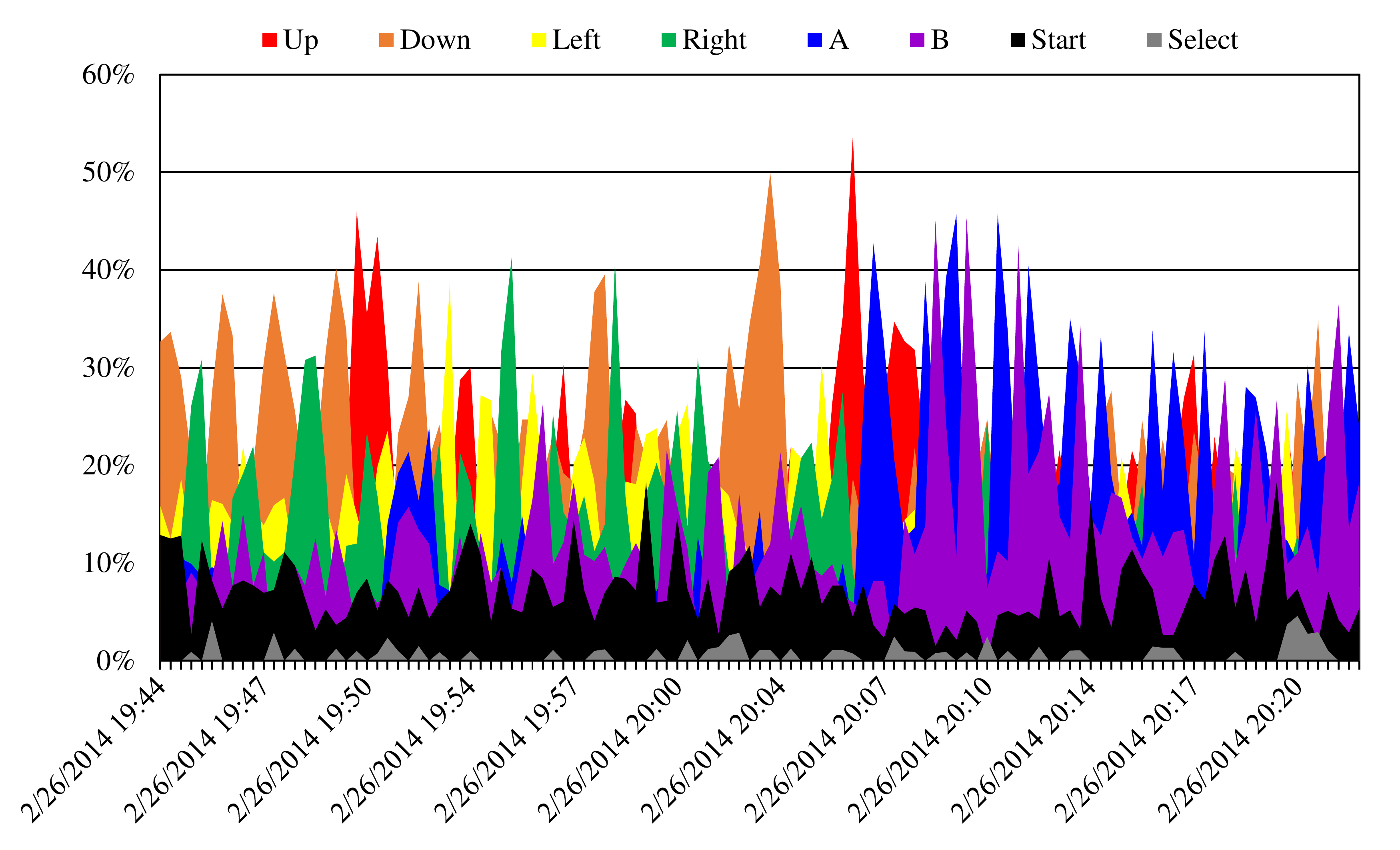}
\vspace{-2 mm}
\caption{Button Frequencies with Context Duration of 20 Seconds}
\label{fig:button_frequencies}
\end{figure}

We chose to use a context duration of 20 seconds for this project. In Figure \ref{fig:button_frequencies}, you can see how the context goal changes over time. The peaks at each point in time indicate which command was most popular in that context. Sometimes it is left, sometimes it is up. For the left half of the graph, we can infer that users are attempting to navigate through the world as the frequencies of A and B are low (i.e. not in a menu). On the right half, the A and B frequencies dramatically increase -- indicating a menu or message dialog has appeared.

\subsection{Feature Set}
We collected information regarding the intervals and total messages sent by each user. This information was not used as features but instead assisted with throwing out impossible anomalies and/or incomplete data points.

The following features were used for clustering and unsupervised learning:
\begin{itemize}
\item $f_1=$  Percent of button inputs that are in the top 1 goal of each context
\item $f_2=$  Percent of button inputs that are in the top 2 goals of each context
\item $f_3=$  Percent of button inputs that are in the top 3 goals of each context
\item $f_4=$  Percent of messages that are spam
\item $f_5=$  Percent of button inputs sent during anarchy mode
\item $f_6=$  Percent of button inputs that are START
\item $f_7=$  Percent of mode inputs that are ANARCHY
\item $f_8=$  Percent of button inputs that are in the bottom 1 goal of each context
\item $f_9=$  Percent of button inputs that are in the bottom 2 goals of each context
\item $f_{10}=$ Percent of button inputs that are in the bottom 3 goals of each context
\end{itemize}

\begin{figure}[h]
\centering
\vspace{-3 mm}
\includegraphics[width=0.6\textwidth]{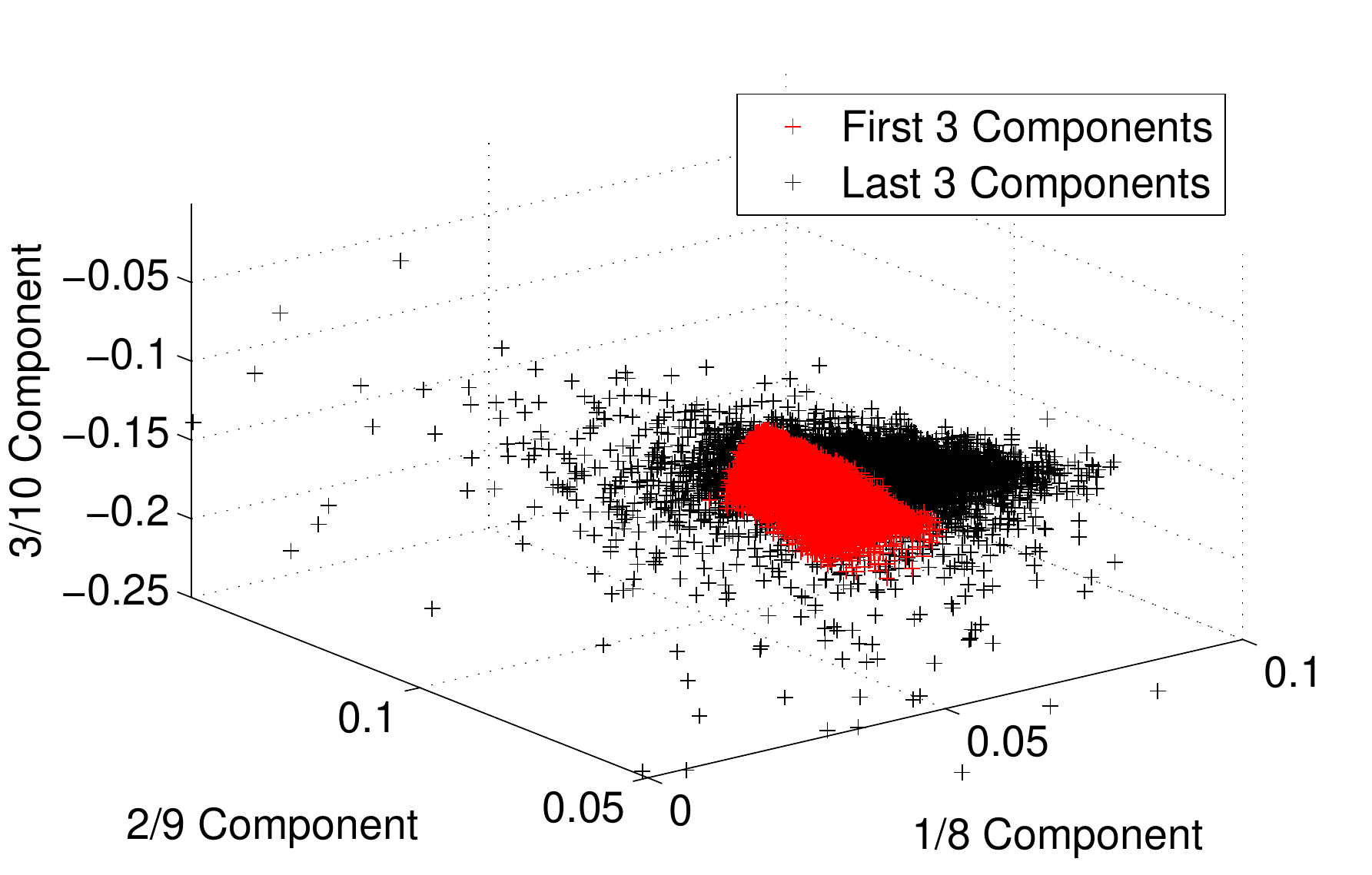}
\vspace{-3 mm}
\caption{First and Last Three Principal Components}
\label{fig:pca}
\end{figure}

Features $f_1,...f_{10}$ were selected after using domain knowledge about the game and the behavior of Internet trolls. Additionally, we used principal component analysis to compare the first three principal components with the last three components. This is shown in Figure \ref{fig:pca}. The components were not used as features.

\begin{equation}\label{eq:svd}
A^T A = VS^2 V^T \text{ and } U=AVS^{-1}
\end{equation}

Because our original dataset consists of 1 million data points, using the standard SVD formula would require significant computational resources. To solve this issue, we used an alternate formation of SVD shown in Equation \eqref{eq:svd}, where $A$ is a matrix containing all 1 million feature vectors. The result of this compression is a reduction from a matrix $A$ (1 million by 10) to matrix $A^T A$ (10 by 10). The latter resulted in a significantly faster running time.

\section{Anomaly Scoring \& Results}\label{sec:algorithms}

We used $k$-means clustering and two distance-based measures to calculate an anomaly score. In particular, we looked at the distance to $k$-nearest neighbor (DKNN) and the sum of the distances to the $k$-nearest neighbors (SKNN). To reduce the algorithm running time, we used a subset of 10,000 feature vectors randomly selected from the original 1 million users. This dataset is provided with the submitted code. All distances were normalized such that the minimum distance had an anomaly score of 0 and the maximum distance had an anomaly score of 100.

\subsection{Distance to $k$-Nearest Neighbor}
\begin{figure}[h]
\centering
\begin{subfigure}[b]{0.49\textwidth}
\includegraphics[width=\textwidth]{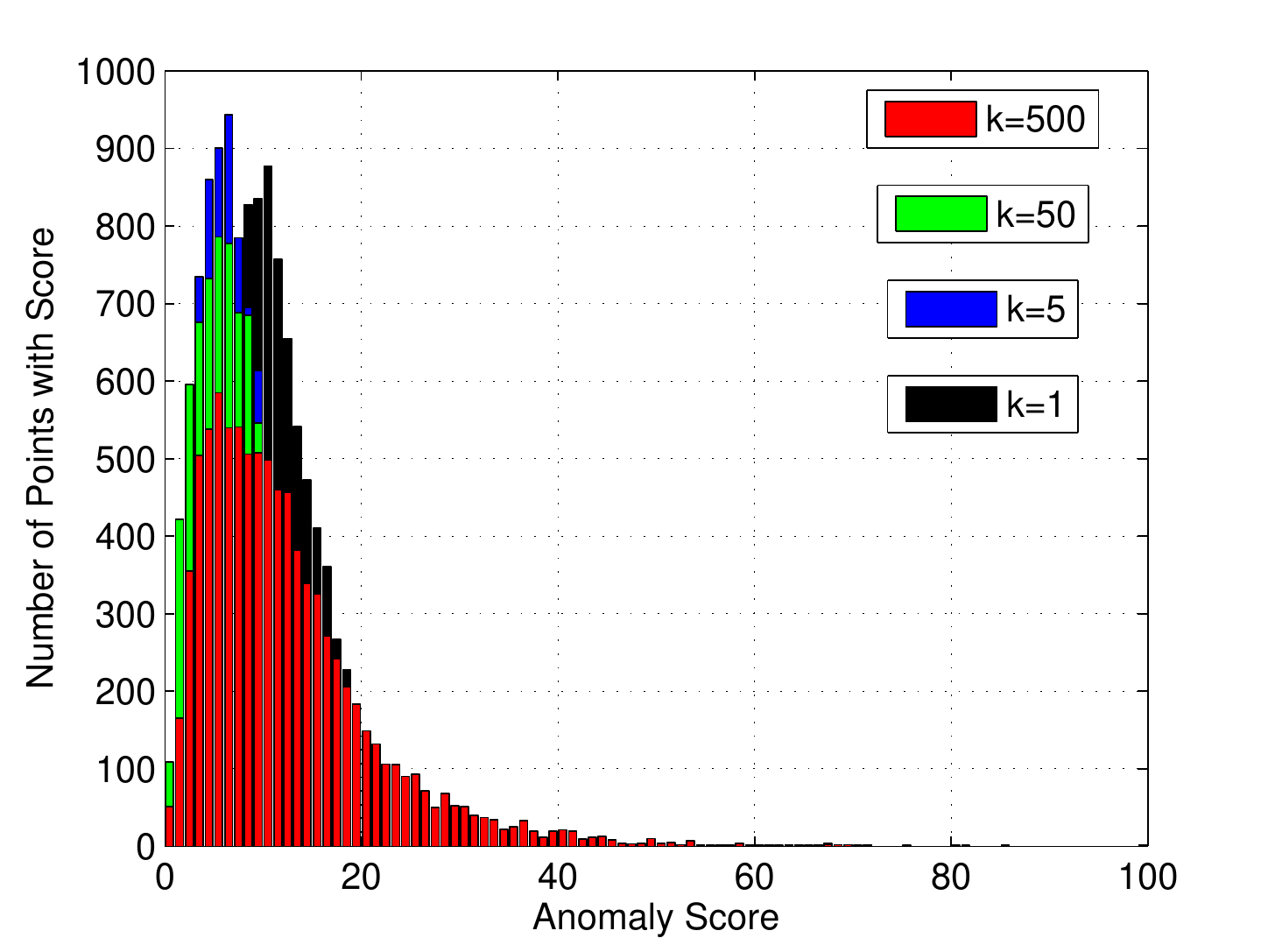}
\caption{Number of Points with Anomaly Score.\\Histogram bucket size is 1.}
\label{fig:dknn-all}
\end{subfigure}
\begin{subfigure}[b]{0.49\textwidth}
\includegraphics[width=\textwidth]{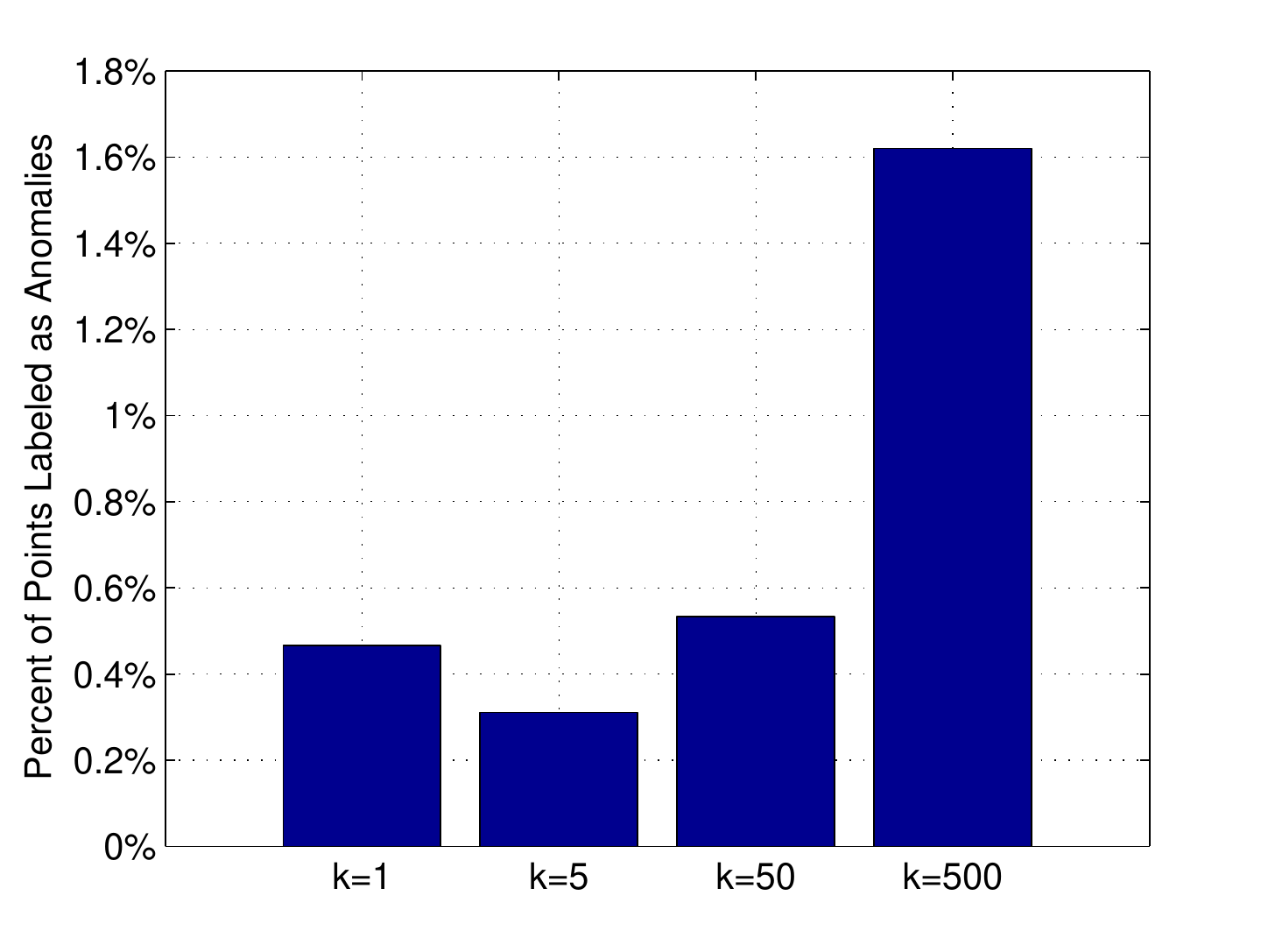}
\caption{Percent of Dataset Labeled as Anomalies Using Anomaly Score Threshold of 40.}
\label{fig:dknn-trollCount}
\end{subfigure}
\caption{Results for Distance to $k$-Nearest Neighbor (DKNN)}
\end{figure}

We used Euclidean distance as the distance metric with $k=1,5,50,500$. Our objective is to minimize the anomaly score of points central to the cluster and maximize the score of those far from the cluster. The results are shown in Figure \ref{fig:dknn-all}.

When $k=1$, it causes the distribution to have the largest median anomaly score. This tells us that $k=1$ is not a good choice. If two anomalies are near each other but far from the main cluster, they will have a low anomaly score. Because of this, we must look at other values of $k$.

With $k=500$, we have a long right tailed distribution (see Figure \ref{fig:dknn-all}). This makes $k=50$ and $k=5$ preferable over $k=500$. After analyzing the effect of $k$, we opted to use $k=5$ as our metric. Given the distribution in Figure \ref{fig:dknn-all}, we used an anomaly score threshold of 40. Any data point with an anomaly score above this value would be labeled as a troll. After labeling the points, the results are shown in Figure \ref{fig:dknn-trollCount}.

\subsection{Sum of Distances to $k$-Nearest Neighbor}
\begin{figure}[h]
\centering
\begin{subfigure}[b]{0.49\textwidth}
\includegraphics[width=\textwidth]{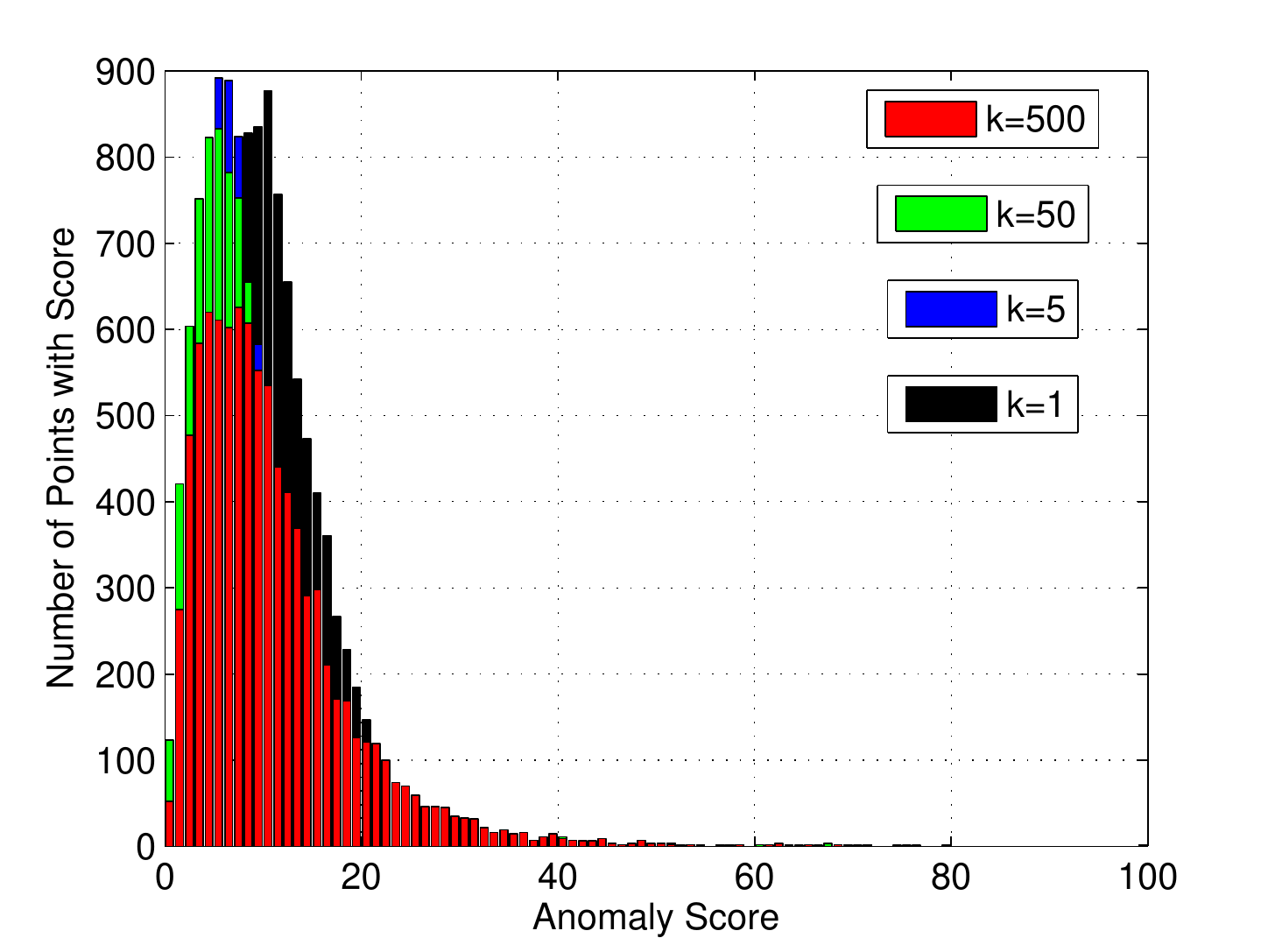}
\caption{Number of Points with Anomaly Score}
\label{fig:sknn_all}
\end{subfigure}
\begin{subfigure}[b]{0.49\textwidth}
\includegraphics[width=\textwidth]{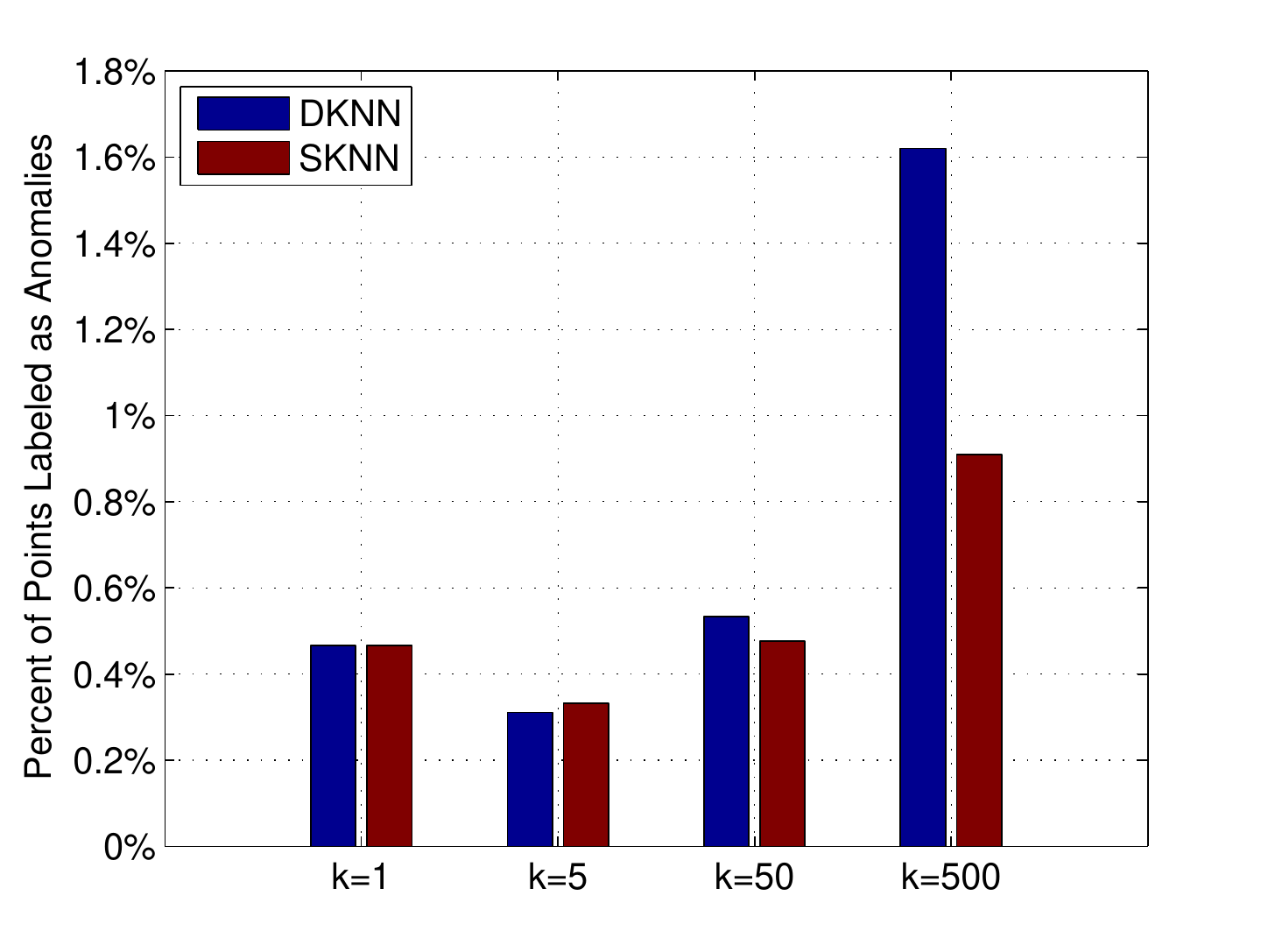}
\caption{Percent of Dataset Labeled as Anomalies}
\label{fig:sknn_trollCount}
\end{subfigure}
\caption{Results for Sum of Distance to $k$-Nearest Neighbor (SKNN)}
\end{figure}

The sum of distances to the $k$-nearest neighbors (SKNN) is similar to DKNN but with one modification: we take the sum of the first to $k^{th}$ nearest neighbor and use this as our distance metric. Results are shown in Figure \ref{fig:sknn_all}.

The SKNN algorithm produces results similar to DKNN. However, when $k=500$, SKNN labels half as many anomalies as DKNN (see Figure \ref{fig:sknn_trollCount}), whereas $k=1,5,50$ produce almost identical results as DKNN.

\subsection{$k$-Means Clustering}
\begin{figure}[h]
\centering
\includegraphics[width=.6\textwidth]{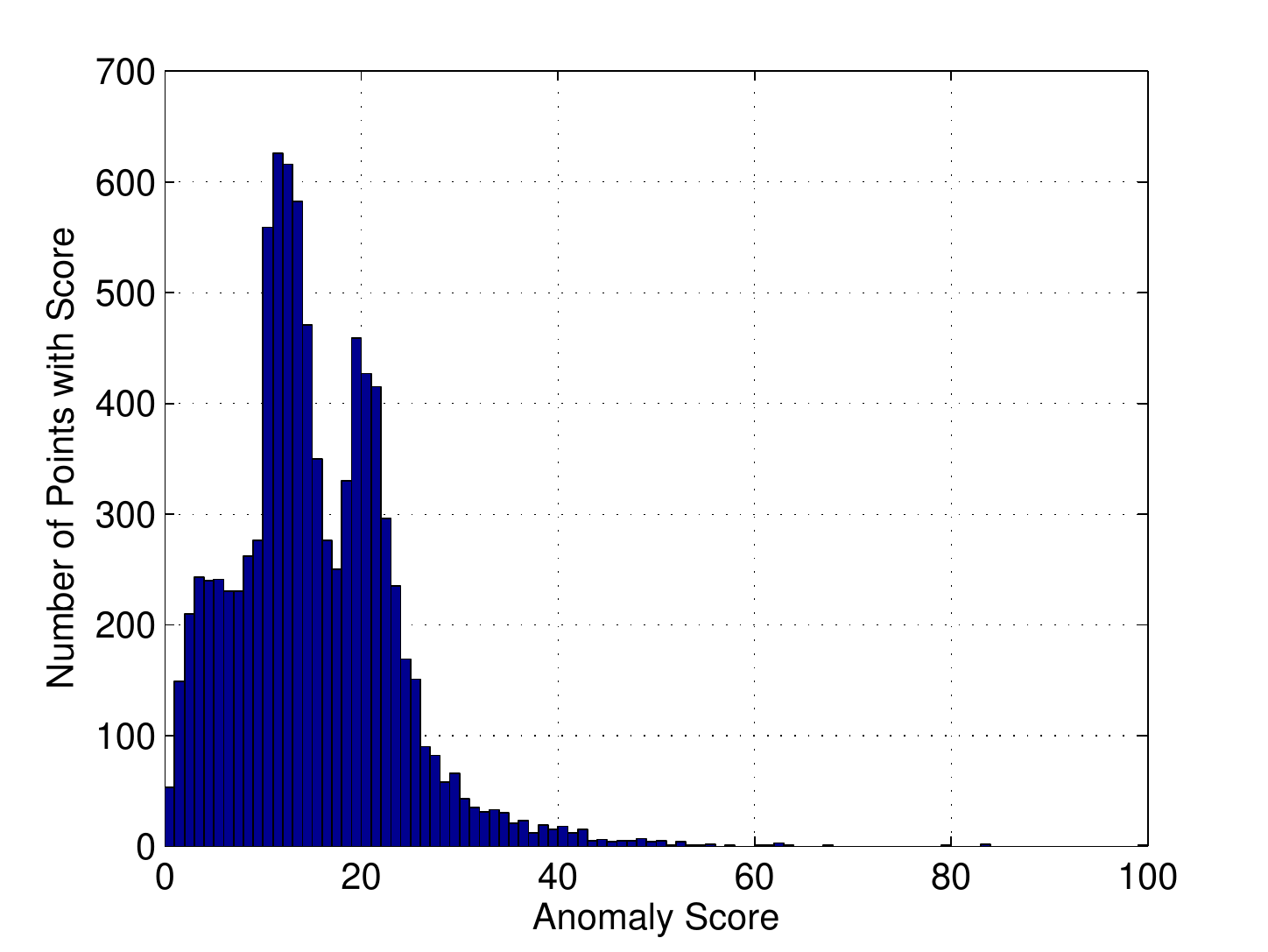}
\caption{$k$-Means Clustering, $k=1$, Number of Points with Anomaly Score}
\label{fig:kmeans-hist}
\end{figure}

In addition to $k$-nearest neighbors, we evaluated the $k$-means clustering algorithm. The value of $k$ was set to $1$. Once the prototype was defined, we used Euclidean distance to estimate each point's distance to the prototype. These distances were normalized and an anomaly score between 0 and 100 was calculated. Higher numbers indicate a higher degree of peculiarity. The algorithm was run using MATLAB and results are shown in Figure \ref{fig:kmeans-hist}.

Also using an anomaly score threshold of 40, $k$-means clustering labeled 107 data points as anomalies which is equivalent to 1.19\% of the dataset.

\subsection{Online Performance}
We apply our anomaly detector to the real-time data stream. Because we cannot obtain features until the context has been generated, there is a maximum delay of 20 seconds until we update our feature vector. We re-cluster the data points once an hour with the updated feature vectors. After re-clustering is complete, we relabel any users who have changed from non-troll to troll and vice versa.

Because this data is unlabeled, we are unable to test the accuracy of this algorithm. However, we have created several test data points (user accounts we own) and have acted as a troll or non-troll. Our online anomaly detector successfully labeled our test data points in the streaming data. Furthermore, we estimate about 1\% of all users are actual trolls.

\section{Discussion} \label{sec:discussion}

\begin{figure}[h]
\centering
\hspace{-7 mm}
\includegraphics[
page=1,
width=0.8\textwidth,
height=\textheight,
keepaspectratio,
trim=40 50 0 20pt,
]{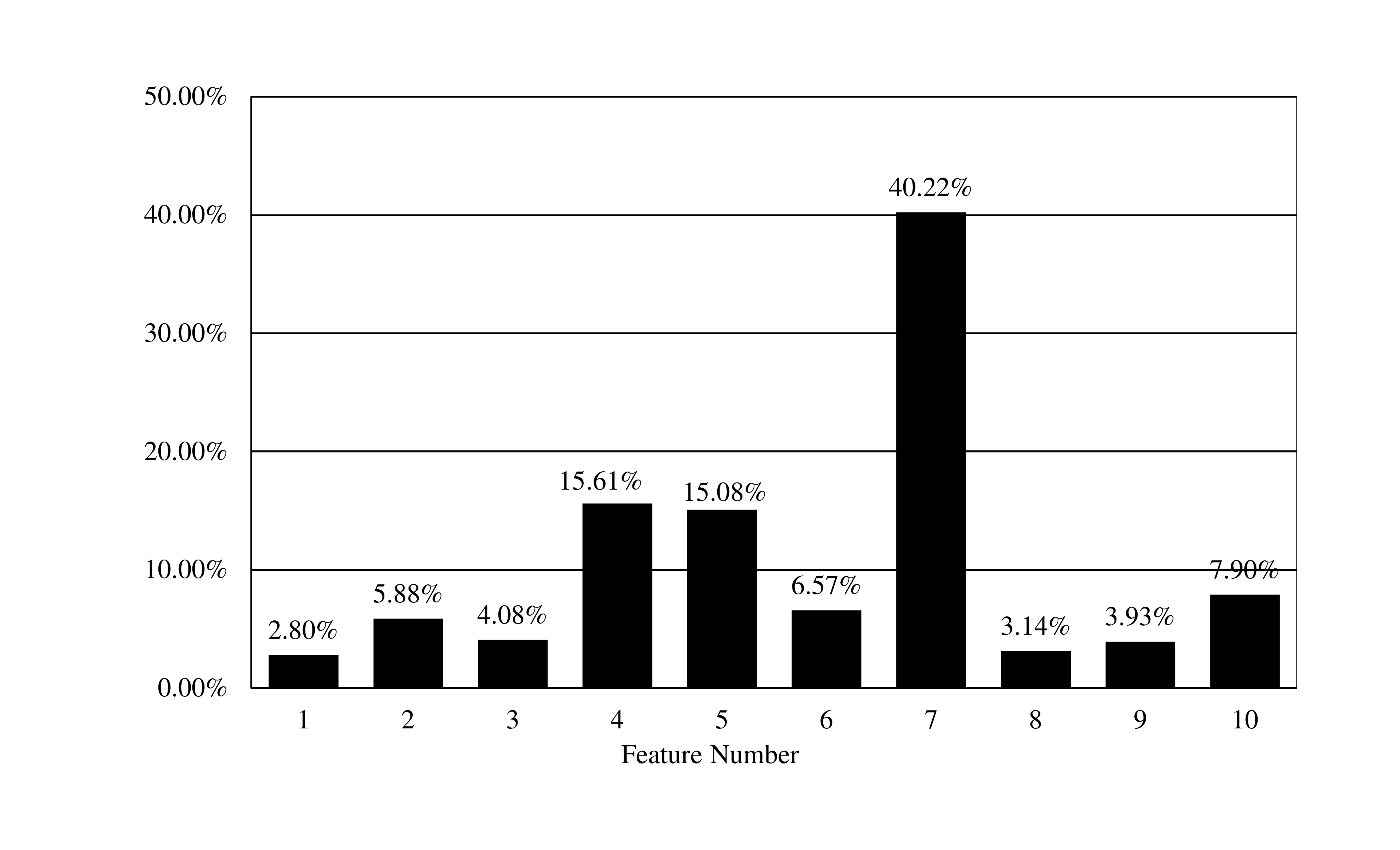}
\caption{Percent of Users Labeled as a Troll by Using Single Feature. This was done by using an arbitrary threshold using domain knowledge.}
\label{fig:featureThreshold}
\end{figure}

Since we do not know the true labels of each data point, the results shown in Section \ref{sec:algorithms} tell us that our definition of an anomaly plays an important role in how many anomalies, or trolls, are labeled. Different trolls will exhibit anomalies in different features and as a result, we cannot look solely at a single feature as a basis for identifying a troll. As shown in Figure \ref{fig:featureThreshold}, if we look at $f_7$ (the number of times a user votes for anarchy) we cannot use this as the sole basis for categorizing users. Simply too many people vote for anarchy, both trolls and non-trolls, and it is very doubtful that 40\% of all users are trolls.

All ten features had a decimal value between 0.0 and 1.0 inclusive. This was generated by analyzing the dataset and corresponds to frequencies. Most features were taken by counting the number of user X's messages that meet the feature's condition divided by the total number of messages sent by user X.

Since a single feature does not work, we must look at more sophisticated techniques such as different distance measures and looked at all ten features simultaneously. When this was done, significantly less users were labeled as anomalies. We estimated about 1\% of all users are true anomalies and Figure \ref{fig:sknn_all} shows how our distance measures were able to classify approximately the same percentage of users as trolls.

\begin{figure}[h]
\centering
\includegraphics[width=0.7\textwidth]{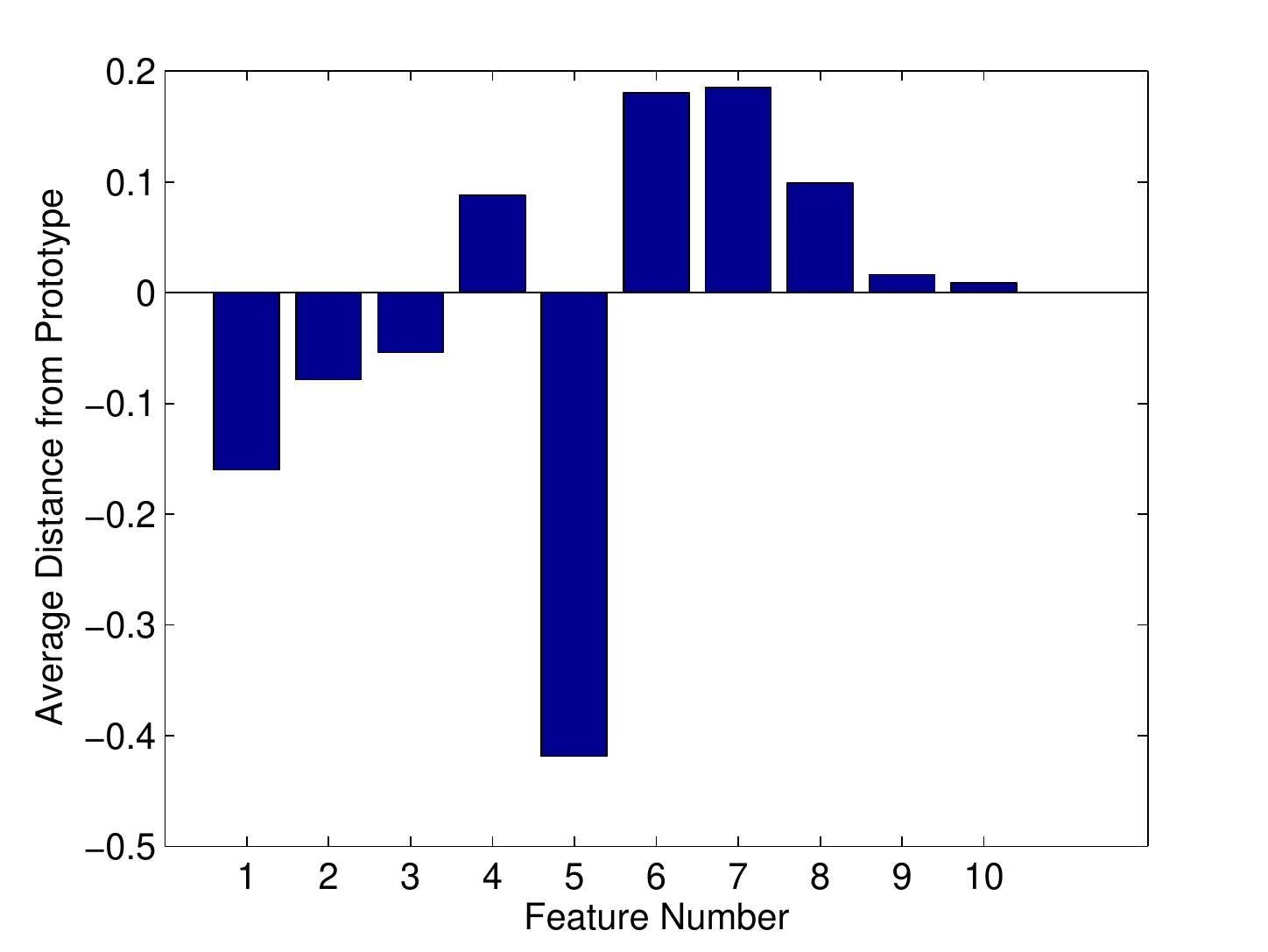}
\caption{Average Distance from Prototype for Troll Features. For all trolls labeled by k-means clustering, the plot shows the average distance from the prototype for each feature.}
\label{fig:trollFeatureDistance}
\end{figure}

We wanted to further understand what differentiates trolls from normal users. To do this, we analyzed the features of trolls labeled by $k$-means clustering. We took these features and calculated how far, on average, did they lie from the $k$-means centroid. The results are shown in Figure \ref{fig:trollFeatureDistance}.

The results tell us that trolls tend not to contribute to the group's goal. Features $f_1, f_2, f_3$ represent the percentage of a user's messages that are in agreement with the group's goal. As shown in Figure \ref{fig:trollFeatureDistance}, trolls are generally not in agreement. Unsurprisingly, trolls tend to submit buttons that are the least frequently submitted in a given context. This is demonstrated by $f_8, f_9, f_{10}$. This tells us that trolls enter buttons that have the least priority and most of the time have negative effects.

Feature 6 counts how often a user pushes START, which is the ``worst" possible move for most contexts. For $f_6$, trolls push START almost 20\% more than the average user. For feature 7, it is expected that trolls vote for anarchy. However, look at Feature 5 -- the number of button inputs that are submitted during anarchy mode. This is surprisingly low and could be explained by the fact that trolls enter the game during democracy, vote for anarchy, then leave once the anarchy has commenced. This is consistent with the definition of a troll who enters a discussion or community with the intention of wreaking havoc and leaves once the job is done.

\section{Conclusion}\label{sec:conclusion}

A lot of the work done here can be extend to anomaly detection in the online setting. During popular interview events, moderators often use an Internet method of collecting questions to ask the speaker. University courses use twitter so students can ask questions anonymously. Constructing contexts by using methods developed in this project can be used to generate contexts during courses and interviews. It would then be possible to filter out irrelevant questions depending on the current lecture slide or segment of the interview.

In \citet{jiang2009detecting}, context is used for mass surveillance videos. The time of day, place, day of the week, and number of people all have an impact on how people act. Modeling this as a context helps machines learn contextual variables – especially in computer vision \citep{jiang2011anomalous}. The work done in this project serves as a starting point for future research in online, context-based communities, and gives researchers new insights as to which areas require greater focus.

\bibliographystyle{iclr}
\bibliography{iclr}

\end{document}